\documentclass{article}

\usepackage{microtype}
\usepackage{graphicx}
\usepackage{booktabs}
\usepackage{amsmath}
\usepackage{float}
\usepackage{caption}
\usepackage{xcolor} 
\usepackage{xurl} 
\usepackage{tabularx}
\usepackage{dblfloatfix}

\usepackage[colorlinks=true, linkcolor=blue, citecolor=blue, urlcolor=blue]{hyperref}
\usepackage[table]{xcolor}
\definecolor{restrgray}{RGB}{240,240,240}
\usepackage{placeins}

\usepackage[preprint]{icml2026}

\icmltitlerunning{The Price of Progress: Price-Performance and the Future of AI}

\begin{document}

\twocolumn[

\icmltitle{The Price of Progress\\
  \vspace{0.3em}
  \large Price Performance and the Future of AI
}

\begin{icmlauthorlist}
\icmlauthor{Hans Gundlach}{mit}
\icmlauthor{Jayson Lynch}{mit}
\icmlauthor{Matthias Mertens}{sloan}
\icmlauthor{Neil Thompson}{mit}
\end{icmlauthorlist}
\icmlaffiliation{mit}{MIT FutureTech, CSAIL, Cambridge, MA, USA}
\icmlaffiliation{sloan}{MIT FutureTech, Sloan, Cambridge, MA, USA}
\icmlcorrespondingauthor{Hans Gundlach}{hansgund@mit.edu}
\icmlcorrespondingauthor{Neil Thompson}{neil\_t@mit.edu}
\icmlkeywords{Machine Learning, AI Progress} 
\vskip 0.3in
]

\printAffiliationsAndNotice{}

\begin{abstract}
\vspace{3mm}

Language models have seen enormous progress on advanced benchmarks in recent years, but much of this progress has only been possible by using more costly models. Benchmarks may therefore present a warped picture of progress in practical capabilities \textit{per dollar}. To remedy this, we use data from Artificial Analysis and Epoch AI to form the largest dataset of current and historical prices to run benchmarks to date. We find that the price for a given level of benchmark performance has decreased remarkably fast, around $5\times$ to $10\times$ per year, for frontier models on knowledge, reasoning, math, and software engineering benchmarks. 
These reductions in the cost of AI inference are due to economic forces, hardware efficiency improvements, and algorithmic efficiency improvements. Isolating out open models to control for competition effects and dividing by hardware price declines, we estimate that algorithmic efficiency progress is around $3\times$ per year. However, at the same time, the price of running frontier models is rising between $3\times$ to $18\times$ per year due to bigger models and larger reasoning demands. Finally, we recommend that evaluators both publicize and take into account the price of benchmarking as an essential part of measuring the real-world impact of AI.

\end{abstract}

\section{Introduction}
A critical dimension of the real-world impact of language models (and AI systems in general) is cost, yet it is often ignored in evaluation discourse. Recent discussion suggests large declines in the price of accessing high LLM performance \citep{appenzeller2024llmflation}. At the upper end, \citet{epoch2025llminferencepricetrends} finds that, controlling for benchmark performance, token prices may be falling by 10--1{,}000$\times$ per year, while \citet{erol2025cost} report more moderate cost-of-pass declines on MATH 500 and AIME 2024 of 24.5$\times$ and 3.23$\times$ per year. Understanding these trends matters for forecasting when models become cost-competitive with labor and for democratizing access to state-of-the-art capabilities.\footnote{Quality-adjusted AI price series are also relevant for economics, e.g., for estimating substitution elasticities between labor and AI inputs.}

In this study, we examine how the price to run LLMs changes for a given performance level and how performance changes with price. This sheds light on key drivers and important heterogeneity underlying trends in AI adoption and ability. We combine benchmark results from the Epoch AI Benchmark Hub \citep{EpochLLMBenchmarkingHub2024} with historical Artificial Analysis price snapshots collected via the Internet Archive, yielding the largest dataset of benchmark-level AI prices we know of (Appendix~\ref{appendix:code and data}). The dataset spans knowledge and reasoning (GPQA-Diamond; GPQA-D), mathematics (OTIS Mock AIME 2024--2025; AIME), and software engineering (SWE-bench Verified; SWE-V).

Using this data, we estimate regressions that recover an exponentially decreasing price trend conditional on performance---a proxy for inference efficiency progress. We use (i) pooled regressions with performance controls and (ii) within-performance-bin regressions; both approaches give consistent qualitative results.

We find overall progress closer to 10$\times$ than 1{,}000$\times$, with estimates generally lower than \citet{epoch2025llminferencepricetrends} and closer to \citet{erol2025cost}. We attribute the gap to three factors: (1) a benchmark- rather than token-level approach that accounts for the rising token usage of recent (especially reasoning) models; (2) a focus on April 2024--November 2025, where we have high-quality price data, rather than earlier periods that may reflect unusually rapid early-stage change; and (3) substantially denser coverage, with nearly an order of magnitude more models per benchmark in our fits.

We also estimate separate trends for open-weight models. Because open-weight models can be run by anyone, they face stronger competitive pressure than closed-weight offerings, so their price declines may better reflect technical change rather than purely economic factors. Using the open-weight trend and adjusting for hardware price progress, we estimate algorithmic progress \footnote{This is a broad definition (see \citet{ho2024algorithmic}). This could include things like better training data, distillation, MoE, etc. } in inference i.e the reduction in the computational operations needed for doing a given task over time \citep{ho2024algorithmic}. This is central to models of AI capability growth \citep{ai2027-about} and to the long-run limits of AI labor scaling when compute cannot grow indefinitely.

Price-performance improvements, however, do not tell the whole story. We also analyze how much benchmark progress is associated with higher inference spending rather than price-independent technical advances. For GPQA-Diamond this effect is particularly strong, roughly half of measured progress is associated with increasing inference prices: holding benchmark cost fixed, performance improves at about half the rate implied by the unconditional time trend. Relatedly, while per-token prices have generally declined, the cost of running frontier-level models has nonetheless risen approximately exponentially (about $3$--$18\times$ per year). This apparent paradox reflects the growing inference required for marginal gains at the frontier, with implications for both the pace of AI development---since evaluation costs may track post-training and reinforcement learning costs---and access to state-of-the-art capabilities.

In addition to establishing these findings we illustrate the current price-performance Pareto-frontier. This helps us identify the importance of open models and mixture of experts in the broader LLM ecosystem. 

Finally, we examine the overall trend in evaluation costs. Some state-of-the-art evaluations can cost thousands of dollars; for example, OpenAI spent about \$3{,}000 per task to run o3-high on ARC-AGI \citep{chollet2024openai_o3_arcagi_pub}. As with frontier models, despite declining conditional price-performance, overall benchmark run costs in our dataset have remained flat or increased to unexpectedly high levels, suggesting that the demand for larger models and greater reasoning has offset (and sometimes overcompensated) per-token efficiency gains.

Finally, we argue that evaluations should be more transparent about computational resource usage. Without such data, it is difficult to distinguish meaningful technical improvements from progress driven primarily by greater compute expenditure.

\section{How Fast Is Benchmark Price-Performance Improving?}\label{sec:How Fast Is Benchmark price-performance Improving}

To measure the trend in benchmark price-performance controlling for quality, we run a variety of regressions on a dataset of benchmarking costs. We first describe our data collection before we turn to our regression results.

\subsection{Data}
 We collect data on input and output token prices over time by gathering Internet Archive data from Artificial Analysis. This data ranges from April 2024 to November 2025 and contains token-level price data from proprietary and open-weight model inference providers (OpenAI, Deepinfra, Cerebras, etc). We collect exclusively the lowest input and output token prices (the full price data across all providers is not available on the Internet Archive). We focus on the cheapest available provider to better control for the price/latency tradeoff and model the lower bound of model inference prices (see \citet{erdil2025inferenceeconomicslanguagemodels}). In some cases, the lowest input and output prices were offered by different providers. However, in the vast majority of cases, the lowest input and output price are offered by the same provider, and when they are not,  the resulting costs differences were negligible.  In addition, we remove datapoints with 0 token cost as these are not reflective of actual inference prices. 

We supplement our dataset with data from Epoch AI's benchmarking hub \citep{EpochLLMBenchmarkingHub2024}. This data includes models' benchmark performance as well as the number of inputs, outputs, reasoning, and cached tokens used to run the benchmark. We compute the benchmark price by multiplying the corresponding tokens by their corresponding prices sourced from Artificial Analysis. 

Prices of running a benchmark may change over time and we treat any such changes as separate data points. As a result, there may be many points for a given model if its price changes frequently. To illustrate: for GPQA-Diamond, which is our largest benchmark sample, we have 138 price data points with 93 unique models; our smallest sample is SWE-bench Verified, which has 21 data points with 19 unique models.  Appendix~\ref{appendix:Details on Dataset Selection, and Preprocessing} provides more details on our data collection procedure.

\subsection{General Regression Approach}

\begin{figure*}[!t]
  \centering
  \includegraphics[width=.70\linewidth]{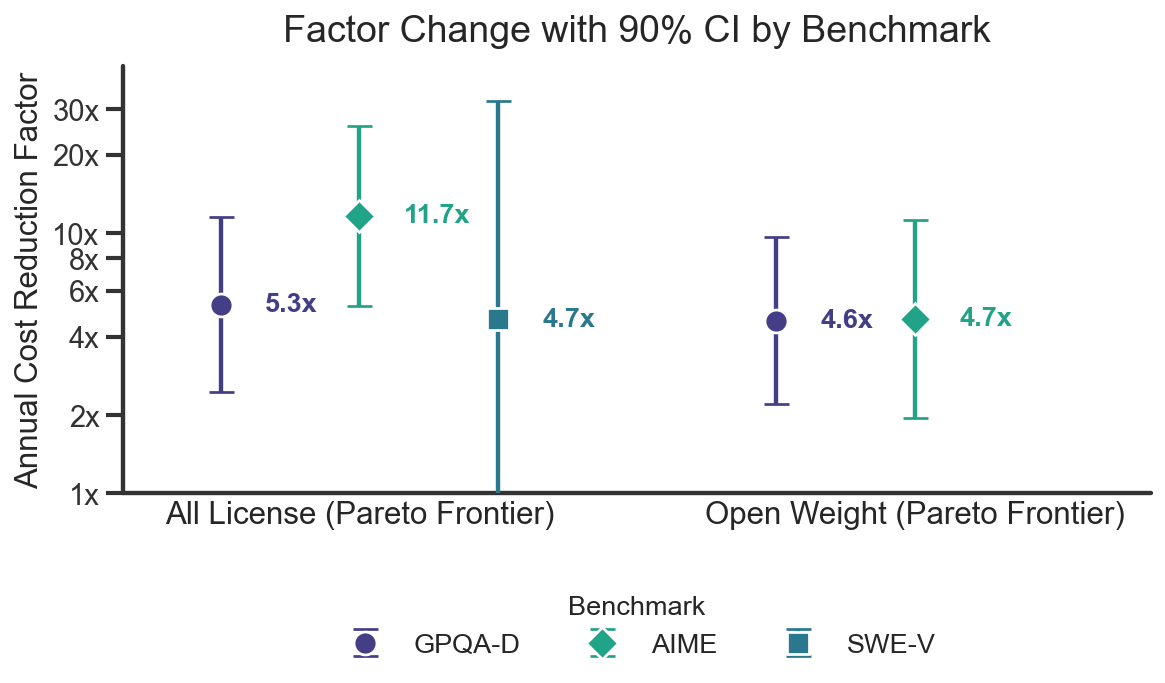}
  \caption{Annual factor change in price controlling for performance. We restrict our analysis to 2024-2025 models on the accuracy-price pareto-frontier (i.e in the Pareto group), and we present separate analyses for all models and only open weight models. Note, we do not have enough open-weight data on SWE-bench Verified to report it (Footnote~\ref{SWE-bench Verified-caveat}). See Table~\ref{tab:regression_results} and Table~\ref{tab:regression_results_adjusted} for more information.}
  \label{fig:growth_rates_combined}
\end{figure*}
\paragraph{Regression specification.}
Our first approach involves running the following regression:

\begin{equation}\label{eq:fit_equation}
   \log(\text{Benchmark Price}_{it}) = \beta_{0} + \beta_{1} \text{logit}(\text{Performance}_i) + \beta_{2} t +  \epsilon_{it},
\end{equation}

where $i$ indexes models, $\text{Performance}_i$ measures benchmark performance of model $i$, $\text{Benchmark Price}_{it}$ is the cost of running model $i$ on the chosen benchmark,  $t$ is time, and $\epsilon_{it}$ is an i.i.d. error term. The coefficient of interest in Eq. \eqref{eq:fit_equation} is $\beta_2$, which measures the rate of log price changes, conditional on benchmark performance. We run separate regressions for each benchmark individually, where $\text{Performance}_i$ measures GPQA-Diamond, SWE-bench Verified, or AIME benchmark scores. Note that we apply a logit-transformation to our performance measure, that is $\text{logit}(Y)=\ln\left(\frac{Y}{1-Y}\right)$, where $Y$ is a reported benchmark score. We choose this transformation for several reasons. First, benchmark performance is bounded between 0 and 1 and increases with the logarithm of inference compute in the non-saturated region \citep {zhang_o1_inference_scaling_laws}. Second, \citet{owen2024predictable} and \citet{ruan2024observational} found that benchmark performance increases logistically with training compute. And since parameters are roughly proportional to inference compute and the square root of training compute \citep{epoch2023tradingoffcomputeintrainingandinference}, we infer that benchmark performance is logistic in log inference compute/price.

\paragraph{Regression samples.}

In our main specification, we estimate the regression for models that are on the Pareto frontier of price-performance at a given time. Specifically, we filter out models that are Pareto dominated by earlier models that have a better benchmark score and a lower price to run a given benchmark. We call models in this group \textbf{Pareto models}. This yields a more relevant sample, as economically-optimizing users will typically choose the cheapest model for a given performance level. In additional analyses (see Table~\ref{tab:regression_results}), we also run the regression for all available models and compare the results.

Finally, we also run separate regressions for open-weight models. Open-weight models can, in principle, be run and modified by anyone. We therefore expect that they will not be priced at a significant markup relative to the necessary GPU resources to run them. As a result, we argue that the trend in open-weight models more accurately measures technical progress rather than economic effects such as increased market pressure, and that the difference between our results for open-weight and proprietary models is (to some extent) informative about non-technological drivers of price changes, such as competition.

One potential concern is that open-weight models might not lie on the overall technical frontier. However, their performance appears to parallel frontier models with a lag of a few months to a year \citep{epoch2024openvsclosedmodelperformance}, so the general technical trend should be approximately similar.\footnote{Almost all the models tested by Epoch AI on SWE-bench Verified are closed-source, so we cannot measure the trend in open-weight price-performance for SWE-bench Verified.\label{SWE-bench Verified-caveat}} The results of our different fits are shown in Fig.~\ref{fig:growth_rates_combined}.

\subsection{Main Results: Frontier Models Are Getting 5-10 Times Cheaper Each Year}


Results for our main specification (Pareto models) are shown in Fig~\ref{fig:growth_rates_combined}. 
We focus first on our main sample. Overall, GPQA-Diamond and AIME benchmark price-performance has increased dramatically, with prices reducing annually by a factor of $5-10\times$ for models on the cost-performance Pareto-frontier.  This is similar to trends in the cost-of-pass for some math benchmarks in \citet{erol2025cost}. For SWE-bench Verified, we have less data, and our estimates feature much larger confidence bands. Nonetheless, the average reduction rate looks similar to our estimates for GPQA-Diamond and AIME. However, our estimates are much lower than those of \citet{epoch2025llminferencepricetrends}. We attribute our lower estimates to three main factors. First, we use a benchmark- rather than a token-level approach, which compensates for the increased number of tokens used in recent models. This is particularly important for reasoning models, which can have a lower per-token cost but a higher overall cost due to generating more tokens. Second, we examine the period from April 2024 to November 2025, where we have high-quality price data. \citet{epoch2025llminferencepricetrends} and related work examine price trends since April 2023 and late 2022, respectively. Therefore, our data do not capture possible price changes during the very initial stages of a new technology. Finally, we include almost an order of magnitude more models per benchmark in our fit than \citet{epoch2025llminferencepricetrends}.

 Tables~\ref{tab:regression_results} and \ref{tab:regression_results_adjusted} report results for all models in our dataset (i.e., including models not on the Pareto frontier). The price-performance decreases are smaller by almost a factor of 2. We think that this likely reflects that our Pareto frontier is defined as optimizing a particular set of benchmarks for cost, but in the real world models are optimized for many different goals (latency, performance for other capabilities, etc.). As such, we do not assume that overall progress is necessarily slower for models in general, just that other model builders are optimizing for other goals.

\subsection{Algorithmic Progress Plays a Big Part in Price Performance Gains}

To get a better sense of the role of algorithmic progress (i.e., technical non-hardware progress) \citep{ho2024algorithmic}, we estimate $\beta_{2}$ (the price decline) for open-weight models only. Additionally, we divide the estimated price trends by the decline in hardware prices as reported in \citet{epoch2024priceperformancehardware}. After these adjustments, the remaining price reduction factor is around \textbf{3$\times$ per year}, which can be interpreted as the contribution of algorithmic progress to declining price. Our estimates for algorithmic progress are more similar to experimental measures of energy efficiency gains like \citep{saad2025intelligence} which find $3.1\times$ gains from 2023-2025 controlling hardware. See  Table~\ref{tab:regression_results_adjusted} for more details on our measurement. 

\begin{figure}[h!]
  \centering
  \includegraphics[width=0.8\linewidth]{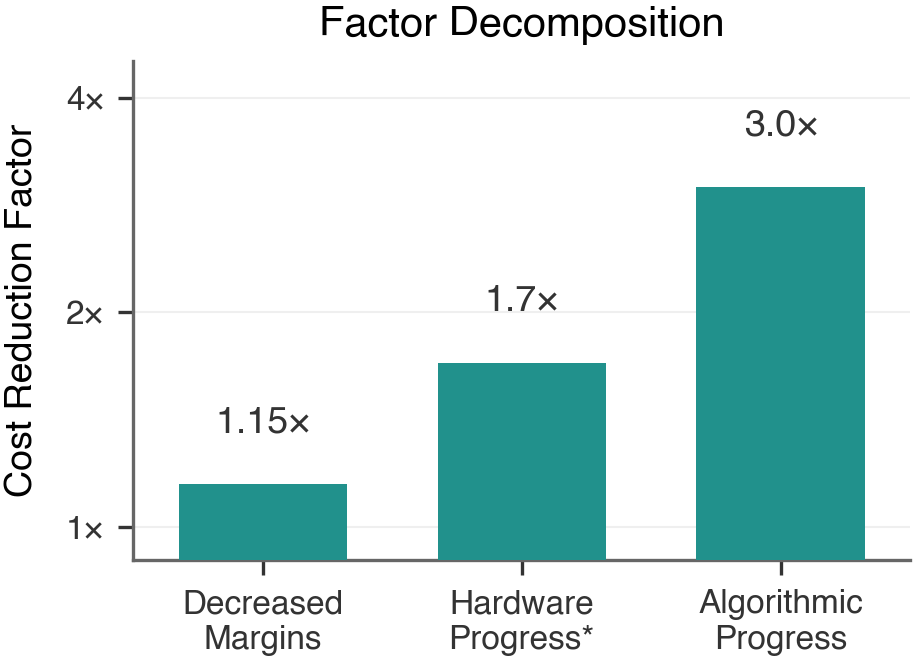}
  \caption{Factor decomposition of decreasing inference prices from 2024-2025 on a log-scale. We estimate that algorithmic progress has played the largest part in overall LLM cost reduction. We estimate hardware progress based on AI GPU trends from \citet{epoch2024priceperformancehardware}.}
  \label{fig:factor_decomposition}
\end{figure}

We also display our factor decomposition in Fig~\ref{fig:factor_decomposition}. We estimate the price performance decrease due to decreased competitive margins by dividing the rate of progress in closed-weight models by the rate of progress in open-weight models (which have much less monopolistic power). We see that our estimate of algorithmic progress is significantly stronger than hardware progress and our estimates of competitive price effects for the time period we measure from 2024-2025.

\subsection{Results by Performance Bins}

\begin{figure*}[!t]
\centering

\begin{minipage}[t]{0.48\textwidth}
\vspace{0pt} 
  \centering
  \includegraphics[width=\linewidth]{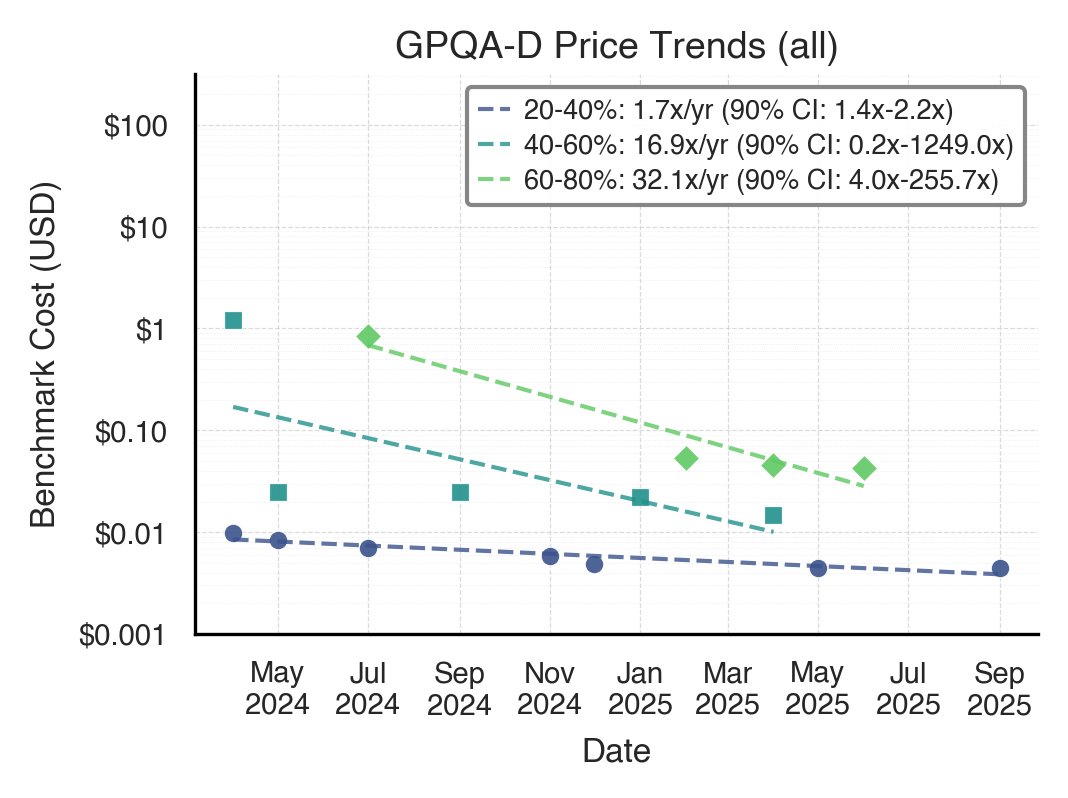}
  \caption{Graph of benchmark price vs time for all models within a fixed GPQA-Diamond range. We don't have a good fit for models in the $40\%-60\%$ range, but include it here for consistency. We suspect that the large drop in overall price in this range is due to increased market competition.}
  \label{fig:all_models}
\end{minipage}
\hfill
\begin{minipage}[t]{0.48\textwidth}
\vspace{0pt} 
  \centering
  \includegraphics[width=\linewidth]{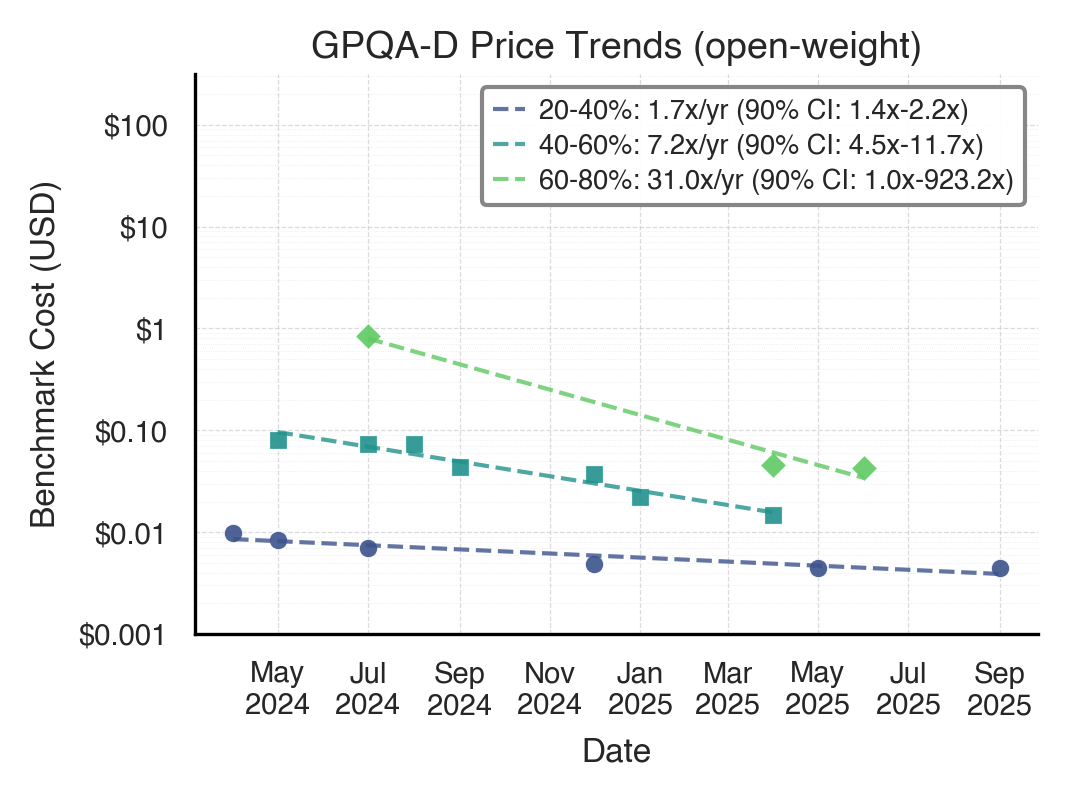}
  \caption{Graph of benchmark price vs time for open weight models within a fixed GPQA-Diamond range. The price for higher quality models is decreasing faster than for lower quality models.}
  \label{fig:open_source}
\end{minipage}

\end{figure*}



In addition to the pooled regression approach above, we also bin models by their benchmark score and estimate Eq.~\eqref{eq:fit_equation} within each bin. Specifically, for each bin we identify, at each point in time, the lowest-priced models, thereby constructing the frontier of capability progress at a given quality level. We then run the regression on the subsample of these frontier models. Fig.~\ref{fig:all_models} depicts the resulting binned trends for all models, while Fig.~\ref{fig:open_source} shows the corresponding results for open-weight models. 

Similar to \citet{epoch2025llminferencepricetrends}, we observe faster trends at the higher-quality frontier. For instance, in Fig.~\ref{fig:all_models}, models in the highest GPQA-Diamond bin declined in price by $31\times$ per year, whereas models in the lowest bin declined by only $1.7\times$ per year. This pattern may indicate that different economic or technical factors drive cost reductions in higher-capability models (e.g., greater use of distillation or Mixture-of-Experts (MoE) architectures).

In addition, we find that the closed-weight model trend is slightly faster than the open-weight model trend. This is particularly pronounced for closed-weight models in the $40\%$--$60\%$ group, where we see a sudden drop in price that is not mirrored in open-weight models, hinting at non-technical competitive effects.

\section{Pareto Frontier Analysis}
\begin{figure*}[t]
  \centering
  \begin{minipage}[t]{0.49\textwidth}
    \vspace{0pt}
    \centering
    \includegraphics[width=\linewidth,height=0.3\textheight,keepaspectratio]{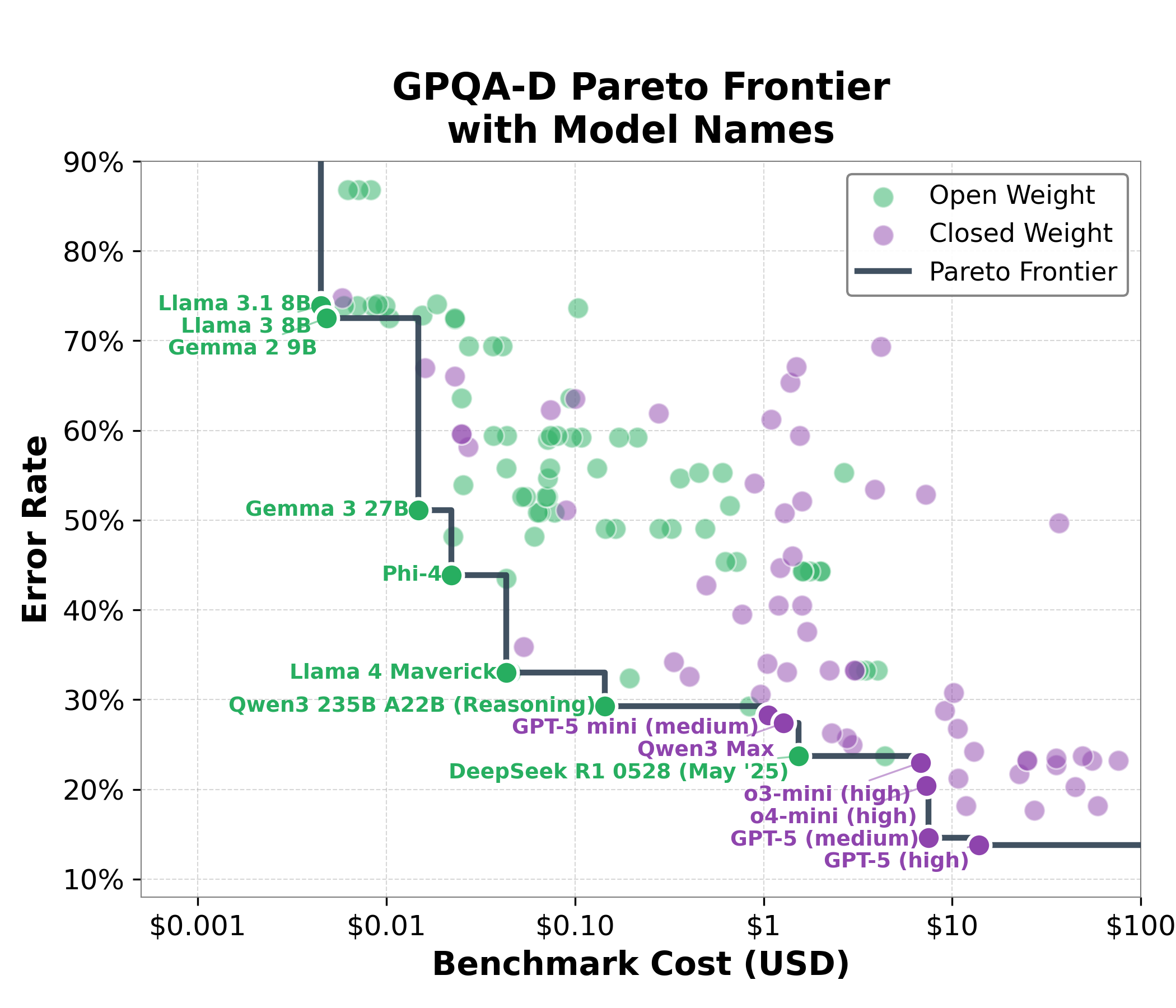}
    \caption{GPQA-Diamond price--performance Pareto frontier (error rate vs.\ benchmark cost) as of November 2025, colored by licensing regime. Open-weight models comprise much of the low- and mid-cost frontier, while closed-weight models more often define the lowest-error end of the frontier.}
    \label{fig:open_closed_pareto}
  \end{minipage}
  \hfill
  \begin{minipage}[t]{0.49\textwidth}
    \vspace{0pt}
    \centering
    \includegraphics[width=\linewidth,height=0.3\textheight,keepaspectratio]{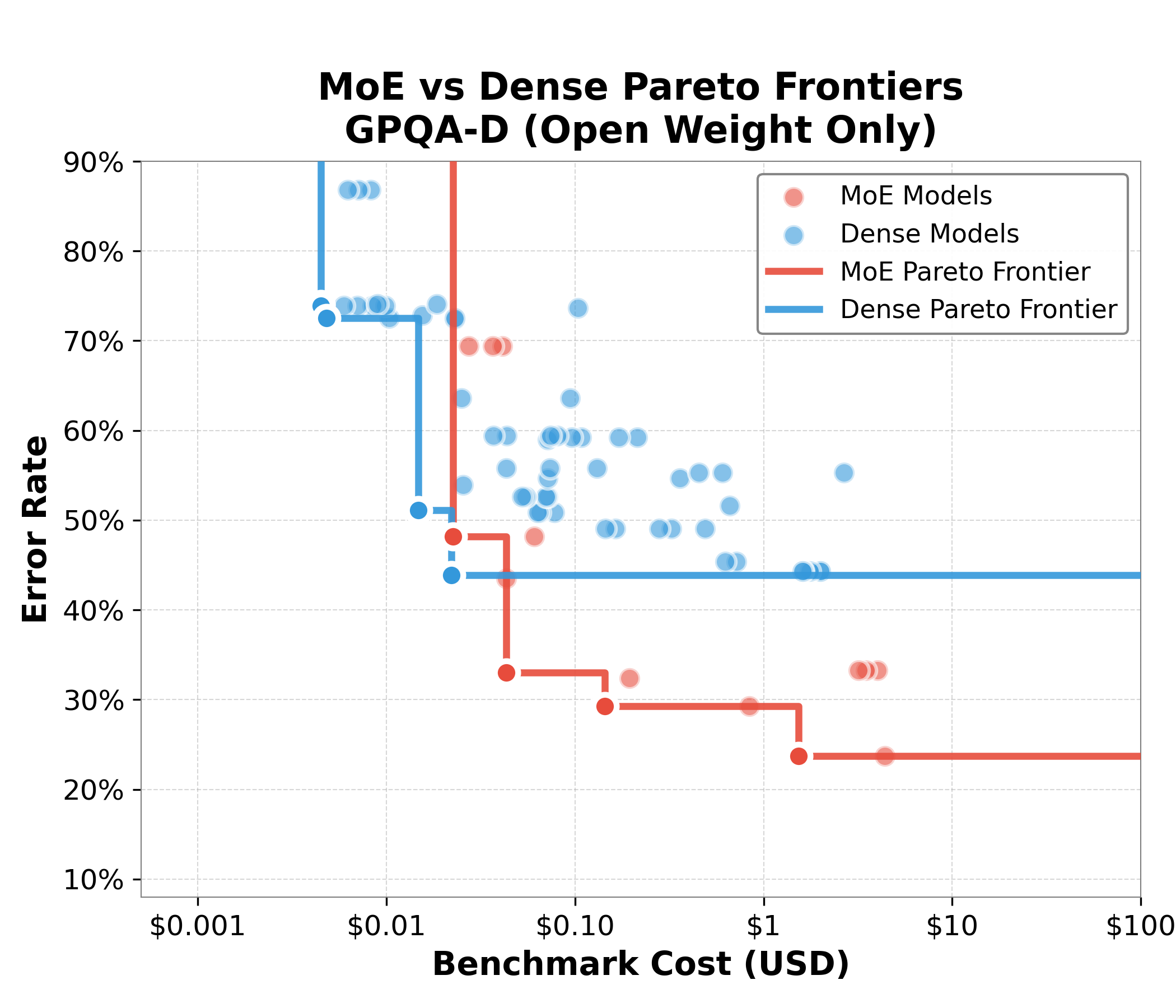}
    \caption{Price--performance frontiers for open-weight models, separated by architecture class (dense vs.\ MoE). Dense models remain most cost-efficient at the very lowest cost levels, while MoE models occupy a growing share of the frontier at lower error rates in our sample.}
    \label{fig:moe_frontier}
  \end{minipage}
\end{figure*}
A natural way to summarize how cost and capability trade off is through the
\emph{price--performance Pareto frontier} at a given time. This helps us view what models and architectures are the most cost-efficient at each capability level for a given benchmark like GPQA-Diamond.

\paragraph{Open vs.\ closed models.}
Figure~\ref{fig:open_closed_pareto} plots GPQA-Diamond error rate against the estimated cost of running GPQA-Diamond, with points colored by whether a model is open-weight or closed-weight. Two patterns stand out.
In contrast to \citet{epoch2024openmodelsreport}, which finds open weight models are one-year behind the best-performing closed models, when we look at the cost-performance Pareto frontier, open models dominate the low- and mid-cost portion of the frontier.
However, the lowest-error end of the frontier is largely defined by closed-weight models, suggesting that proprietary systems disproportionately push the peak benchmark performance currently attainable.

\paragraph{MoE vs.\ dense among open-weight models.}
To examine the role of architecture within open-weight models, Figure~\ref{fig:moe_frontier} restricts the sample to open
models and separates the frontier by model class: dense versus mixture-of-experts (MoE). We plot the Pareto frontier formed
by dense open models alongside the frontier formed by MoE open models. The resulting picture suggests that MoE models
occupy much of the frontier at moderate and high cost levels and appear increasingly advantageous at lower error rates,
while dense models remain more cost-efficient at the lowest accuracy levels. In our sample, dense models also appear to
plateau around a $40\%$ GPQA-Diamond error rate. We theorize that each of the experts in a mixture of experts models needs a critical size; if dividing a model causes experts to be below this size, it is not useful to do so. In addition, MoE models generally have routing, load-balancing, and communication costs that hinder the efficiency of small-scale usage.

However, these comparisons are descriptive and should be interpreted cautiously. Architecture is confounded with other design choices:
MoE models in our sample are often newer and may incorporate additional innovations (e.g., improved data, training recipes,
post-training, or inference-time policies) beyond the MoE routing mechanism itself. As a result, we do not interpret the
MoE--dense gap as causal, but rather as suggestive evidence about where different model families lie on the observed cost--performance
frontier.

\section{AI Progress Slows When Inference Prices Are Held Constant}

\begin{figure*}[t]
\centering
\begin{minipage}[t]{0.49\textwidth}
\vspace{0pt}
  \centering
  \includegraphics[width=\linewidth]{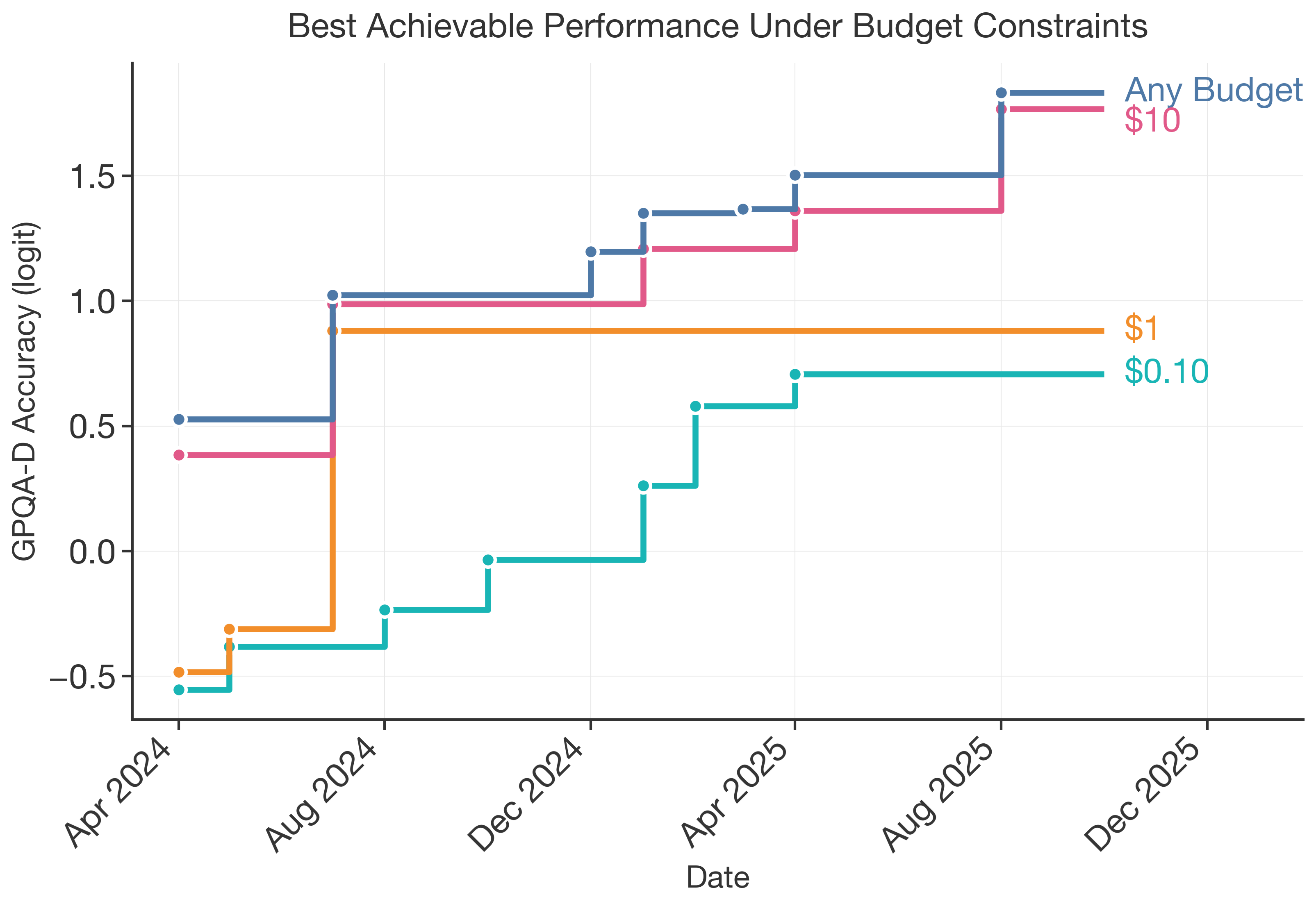}
  \caption{Progress of best performing models on GPQA-Diamond below a given benchmark price level. Dark blue step function depicts best performance without any price restriction. Notably, models at lower price ranges are seeing slower progress.  }
  \label{fig:gpqa_budget_epoch}
\end{minipage}
\hfill
\begin{minipage}[t]{0.47\textwidth}
\vspace{0pt}
  \centering
  \includegraphics[width=0.8\linewidth]{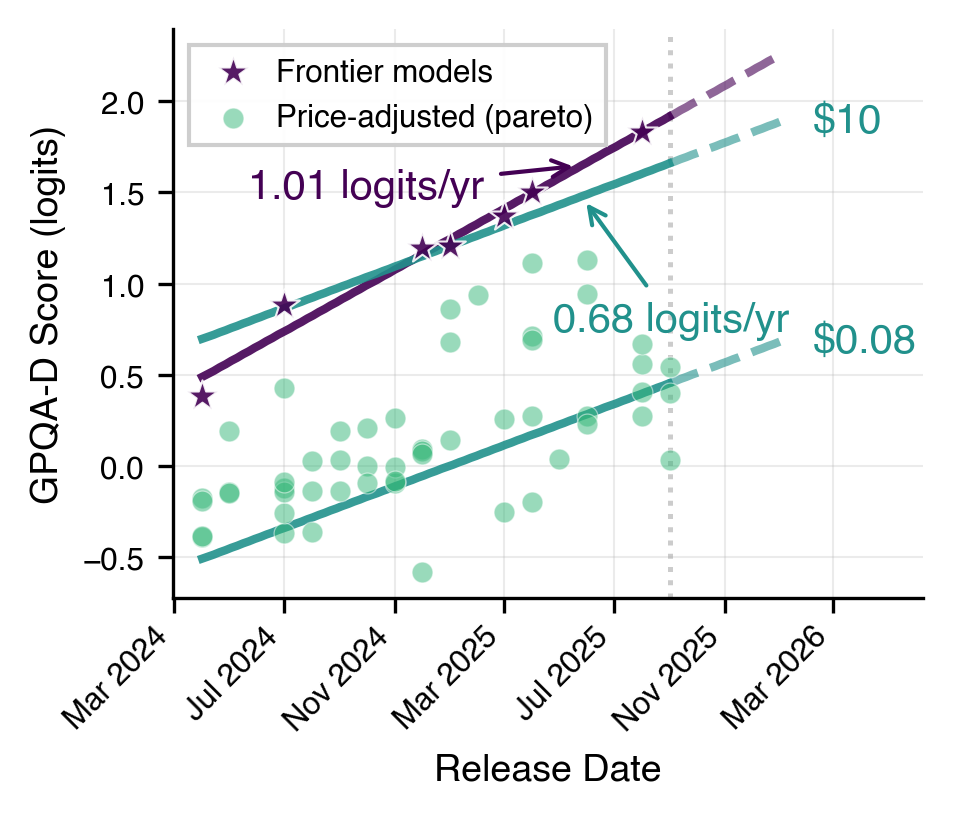}
  \caption{GPQA-Diamond progress with and without cost controls. 
 Blue points show trends in Pareto models after residualizing/adjusting with respect to $\log(\text{BenchmarkPrice})$. $\$0.08$ represents the median price of models that were ever on the Pareto frontier. Purple points are frontier logit benchmark scores (not-residualized). The gap between trends indicates that a large share of measured progress is associated with higher inference prices for GPQA-Diamond performance as in Fig~\ref{fig:gpqa_budget_epoch}.}
  \label{fig:gpqa_cost_adjusted_progress}
\end{minipage}
\end{figure*}

We have focused up to this point on looking at price trends at fixed benchmark score but we can equally look at trends in benchmark score with a fixed price. 
Benchmark performance has improved rapidly in recent years, but these gains have not occurred at a fixed level of
inference expenditure. In many practical settings, including deployment of agentic systems, inference costs impose
binding constraints on usage and diffusion. It is therefore important to separate (i) capability progress that reflects
price-independent improvements (e.g., algorithmic and systems efficiency) from (ii) progress that is primarily achieved by spending more at inference time. We do this in two ways. 
First,  we look at progress in frontier models with benchmark prices under a given level. This is shown in Fig~\ref{fig:gpqa_budget_epoch} as well as in Appendix~\ref{appendix:added_budget_graphs}. In the second method, we run regressions on selected models with and without controlling for benchmarking price. This is represented in Fig~\ref{fig:gpqa_cost_adjusted_progress} and Table~\ref{tab:regression_results_adjusted}. Both methods, yield broadly similar results, showing benchmark progress slows when limiting total inference expenditure. Historical frontier GPQA-Diamond progress is strongly influenced by increasing benchmarking/inference costs while frontier SWE-bench Verified and AIME progress are significantly less influenced by cost increases.
One possible explanation is that AIME and SWE-bench Verified have benefited significantly from RL enhancements, which don't directly increase inference costs.

\subsection{Estimating the Contribution of Higher Inference Prices}
We quantify how much measured benchmark progress is associated with rising inference prices by comparing two
\emph{nested} regression specifications. The first estimates the unconditional time trend in performance:
\begin{equation}
\label{eq:progress_unconditional}
\text{logit}\!\left(\text{Performance}_{it}\right) = \alpha + \beta_{t}\, t + \varepsilon_{it},
\end{equation}
and the second controls for benchmark cost:

\begin{equation}
\label{eq:progress_cost_controlled}
\begin{aligned}
\text{logit}\!\left(\text{Performance}_{it}\right)
&= \alpha + \beta_{t}^{\,\text{(cost)}}\, t \\
&\quad + \beta_{p}\, \log\!\left(\text{BenchmarkPrice}_{it}\right)
+ \varepsilon_{it}.
\end{aligned}
\end{equation}
Here, $\text{Performance}_{it}$ is the benchmark score of model $i$ (e.g., GPQA-Diamond) and
$\text{BenchmarkPrice}_{it}$ is the estimated cost of running that benchmark on model $i$ at time $t$.

The key object of interest is the change in the estimated time coefficient when we add cost controls,
$\beta_{t} - \beta_{t}^{\,\text{(cost)}}$. Intuitively, $\beta_{t}$ captures how quickly benchmark performance improves over time
in the observed model population, whereas $\beta_{t}^{\,\text{(cost)}}$ measures the time trend \emph{holding benchmark cost fixed}.
A large gap between these two coefficients indicates that much of apparent ``progress over time'' is associated with
using more expensive inference.\footnote{Conceptually, this approach is similar to \citet{mertens2026there} who isolate the impact of time effects (progress) on LLM performance by controlling for model training compute.}

\paragraph{Samples.}
We estimate Eqs.~\eqref{eq:progress_unconditional}--\eqref{eq:progress_cost_controlled} on (i) all models in our dataset and
(ii) the price--performance Pareto frontier (the Pareto Sample), defined as the subset of models that are not dominated in the
(cost, performance) plane at a given time $t$ \footnote{A model is dominated if there exists another model that achieves at least as high a
benchmark score at no greater cost, with strict improvement in at least one dimension.} and (iii) the best performing Frontier models at any given time. The Pareto sample focuses
attention on the best performance achievable at each cost level and is therefore most relevant for economically optimizing users.

\subsection{Much of GPQA-Diamond Progress can be Attributed to Higher Prices}
As shown in Fig~\ref{fig:gpqa_budget_epoch}, progress for models with bounded benchmark cost has plateaued for GPQA-Diamond. However, repeating the same analysis in Appendix~\ref{appendix:added_budget_graphs}, we show that SWE-bench Verified and AIME have similar benchmark growth regardless of benchmark/inference budget.

Figure~\ref{fig:gpqa_cost_adjusted_progress} illustrates the effect of increasing costs on GPQA-Diamond. The blue series plots the corresponding
\emph{cost-adjusted} scores, obtained by residualizing/adjusting performance with respect to $\log(\text{BenchmarkPrice}_{it})$ for models in our Pareto-group
(i.e., removing the component of performance statistically associated with higher evaluation cost).  The Frontier trend is $48\%$ higher than the Pareto-adjusted group for GPQA-Diamond. We see similar but smaller effects using the same analysis on SWE-bench Verified and AIME. When doing the same comparison for SWE-bench Verified and AIME, we see the frontier is progressing $12\%$ and $14\%$ faster than the cost-adjusted Pareto models, respectively.

\section{Inference Prices for Frontier Models Are Increasing Rapidly}

\begin{figure}[h!]
    \centering
    \includegraphics[width=\linewidth]{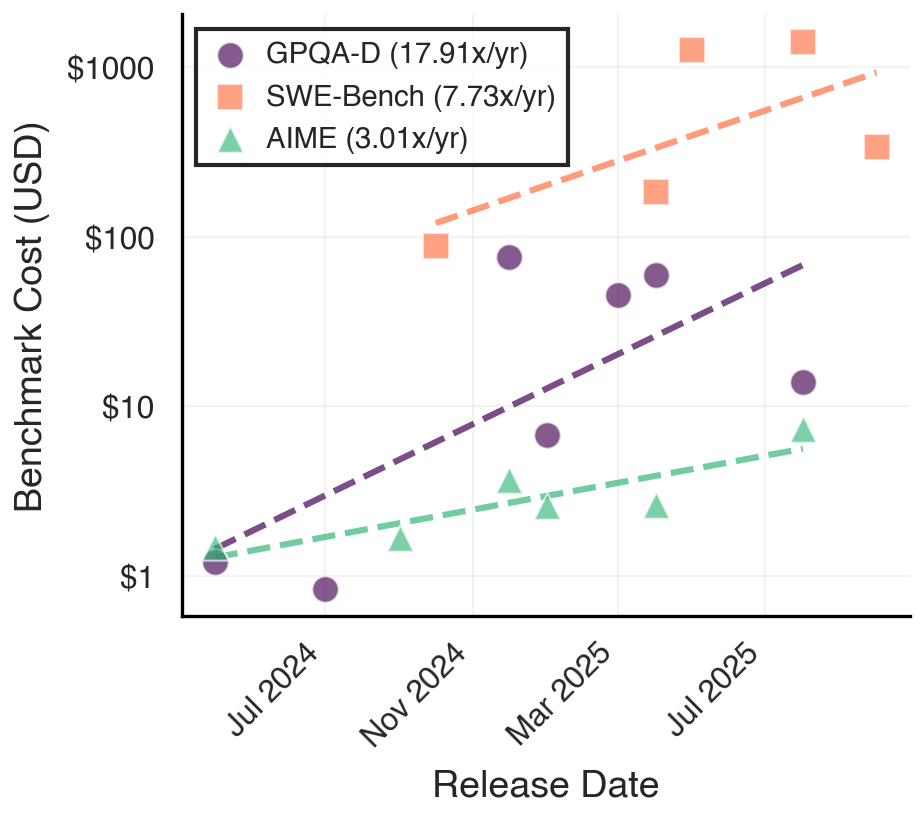}
    \caption{Estimated benchmark/inference cost required to reach frontier (record) performance on GPQA-Diamond, SWE-bench Verified, and AIME over time.}
    \label{fig:frontier_bench_price}
\end{figure}

Figure~\ref{fig:frontier_bench_price} plots the estimated inference cost of running the best-performing available model on three benchmarks (GPQA-Diamond, SWE-bench Verified, and AIME) over time. While per-token inference prices generally decline over time, the cost of evaluating (and therefore using) frontier-level models on these benchmarks have nonetheless increased at an approximately exponential rate, on the order of $3$--$18\times$ per year.

These trends are not contradictory. Per-token inference can become cheaper while the cost of \emph{frontier performance} rises if achieving marginal performance gains requires substantially more inference. For example, if inference prices fall by $10\times$ but improving performance by $1\%$ requires $100\times$ more inference, then the total cost of running the state-of-the-art model at that new level increases by $10\times$.

This matters for two reasons. First, evaluation costs are plausibly tied to the cost of reinforcement learning (RL) and post-training procedures, which require large volumes of model-generated rollouts on increasingly complex tasks and data. If evaluations become substantially more expensive, then iteration slows: obtaining reliable feedback on model changes and running RL rollouts becomes more costly and time-consuming.

\begin{figure*}[!t]\label{fig:benchmark_costs}
\centering
\begin{minipage}[t]{0.48\textwidth}
\vspace{0pt}
  \centering
  \includegraphics[width=\linewidth]{ 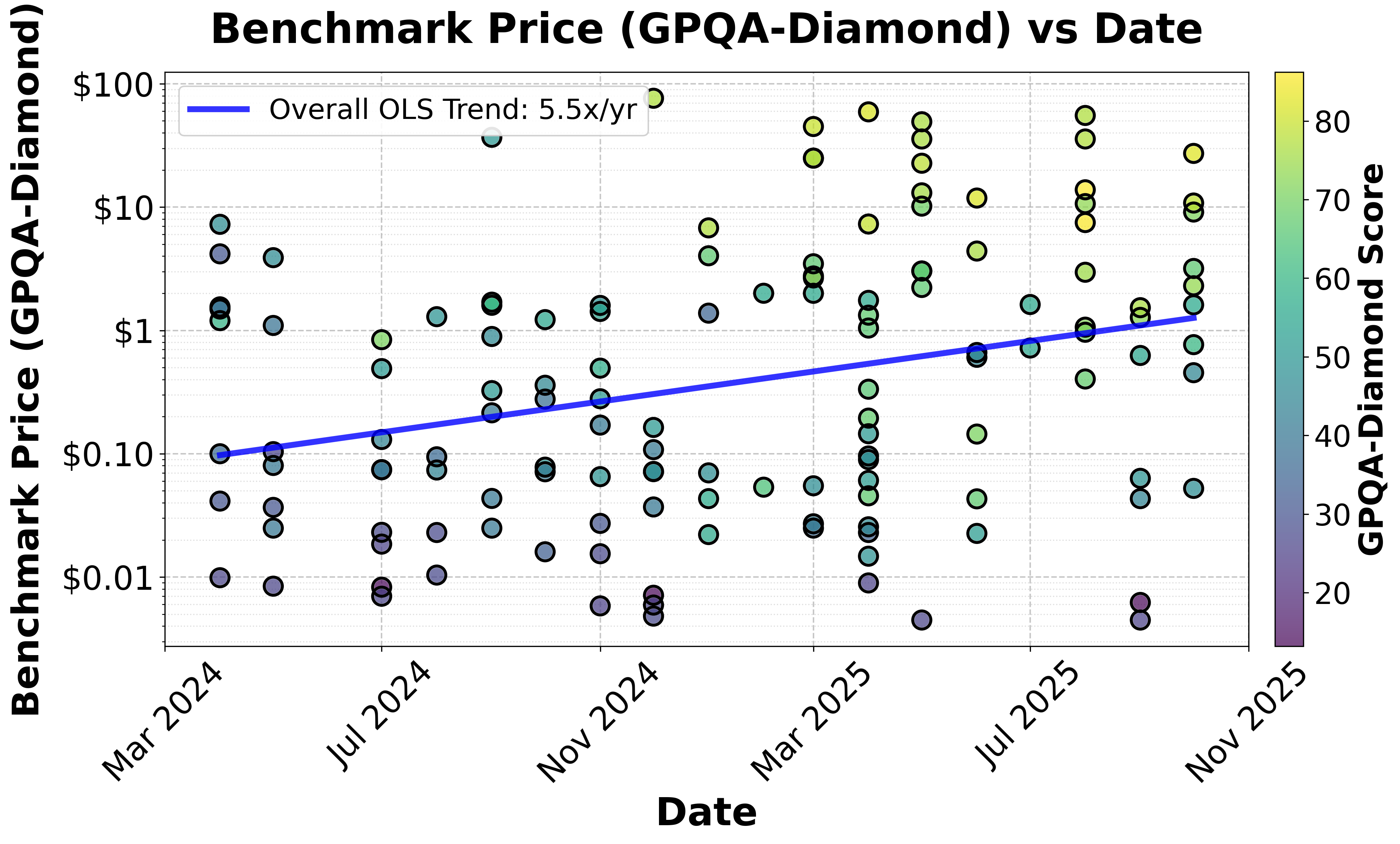}
  \caption{Price to run GPQA-Diamond benchmark. Prices based on Epoch-AI benchmark data and Artificial Analysis Prices. Overall, benchmark prices in our dataset have increased despite a dramatic fall in model price-performance.}
  \label{fig:GPQA-Diamond-Price}
\end{minipage}%
\hfill
\begin{minipage}[t]{0.48\textwidth}
\vspace{0pt}
  \centering
  \includegraphics[width=\linewidth]{ 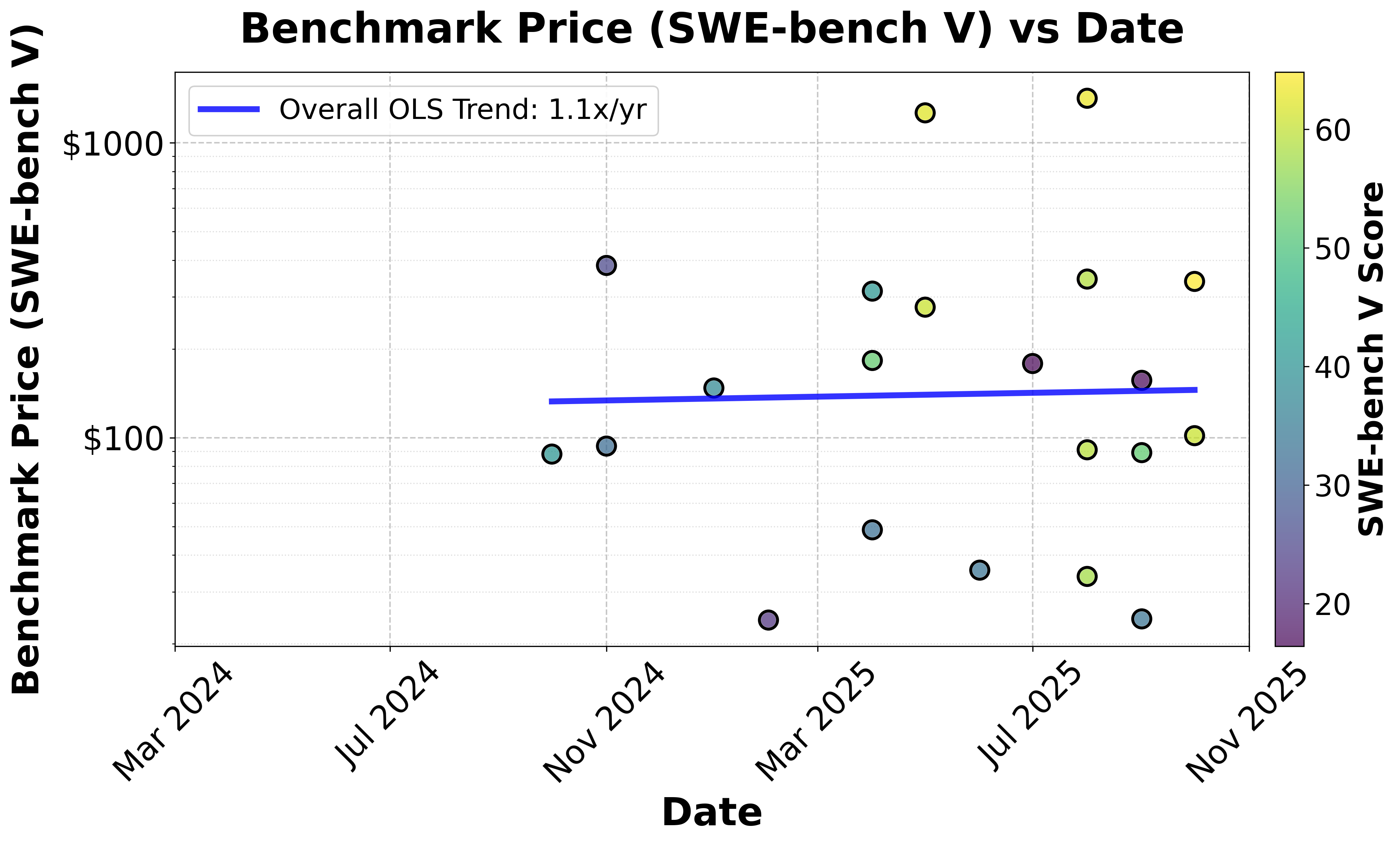}
  \caption{Price to run SWE-bench Verified. Prices based on Epoch-AI benchmark data and Artificial Analysis Prices. Similar to Fig~\ref{fig:GPQA-Diamond-Price}, benchmark prices have increased. In addition, the price to run SWE-bench Verified for some models is now in the thousands of dollars.}
  \label{fig:swe_cost_v_time}
\end{minipage}
\end{figure*} 
Second, evaluation costs are also indicative of the cost of deploying frontier models in real-world settings. If frontier-level inference becomes expensive faster than prices fall, pushing the frontier may become less economically attractive for broad deployment. This could also increase inequality in access to capabilities: today many consumers can use the best available models, but if frontier inference costs continue to rise, access may concentrate among a smaller set of high-budget users and organizations.

\section{Inference Costs Are Falling, but Overall Benchmarking Costs are Increasing}

We close our paper by examining the overall cost of benchmarking AI models regardless of frontier performance.  Benchmarking is a fundamental part of AI research and development. Large benchmarking costs pose a challenge to maintaining a healthy evaluation ecosystem. We take the previous price estimates of benchmarking and look at the overall trend regardless of performance. We find that, despite the dramatic fall in the price for a given performance level previously discussed, benchmarking costs have overall stayed constant or increased (see Figs.~\ref{fig:GPQA-Diamond-Price} and \ref{fig:swe_cost_v_time}). We believe that this reflects the demand for much higher quality models, which are larger and use much longer reasoning traces. 

For SWE-bench Verified, average evaluation prices have remained roughly constant. However, beneath the stable overall trend lies considerable model-specific variation in evaluation prices: The cost of running SWE-bench Verified on some models has risen to thousands of dollars. Overall, increasing evaluation prices can be taxing for small academic research groups, and this problem is compounded if many different models must be run or if multiple benchmarks are involved in a study. Our analysis does not take into account new, longer contexts and agential benchmarks, which are becoming prohibitively expensive. For instance, {$\infty$\kern0.15em BENCH} spent 5,000 dollars evaluating long-context abilities on GPT-4 \citep{zhang2024infty}. Such costs make independent and academic benchmarking prohibitive and have led to the rise of private firms like Artificial Analysis, which have the capital necessary to do cutting-edge evaluations on many models. These companies are valuable resources but only share some information with the public and keep other information, like benchmark costs and historical prices, either hidden or not easily accessible.

\section{Conclusions and Recommendations for Future Evaluations}

At first glance, our analysis seems to point in different directions. The overall price for accessing a
given level of LLM performance has dropped significantly, by 5$\times$ to 10$\times$ per year, although still
substantially less than the reported 1000$\times$ upper bound by \citet{epoch2025llminferencepricetrends}. However, like \citet{epoch2025llminferencepricetrends}, we find much larger price declines for higher-performance models---almost 32$\times$ per year.
For the least performant models, by contrast, we see much smaller price declines, around 1.7$\times$ per year,
close to the estimates for energy efficiency improvements in AI models overall \citep{saad2025intelligence}.
In terms of benchmarks, progress looks similar across GPQA-Diamond and AIME, but the data for SWE-bench Verified is so
limited that our confidence bounds are large enough to be consistent with there having been no progress at all.

Despite these large price decreases, the benchmarking costs of models in our dataset have remained constant or increased as the demand for quality has risen, leading to larger models and longer reasoning. This is a particularly strong tension in frontier models: while a fixed capability level becomes dramatically cheaper over time, achieving frontier performance continues to demand higher frontier expenditure. Further, for some benchmarks a significant portion of apparent benchmark progress can be attributed to using more expensive models.

Taken together, these findings suggest that focusing on benchmark scores alone does not provide a full view of the nuanced nature of AI progress. A model that substantially improves benchmark performance but only by consuming dramatically more computational resources represents a smaller advance in practical capability than headline scores suggest. Hence,  we must take into account price if we want to understand progress.

We therefore call for evaluators and researchers to publish price and resource data on evaluations more
widely, so that the evaluation community can examine capability progress in light of real-world economic
constraints. Given the limited data available, it is hard to know what real-world progress has been made
in SWE-bench or in any benchmark. Has math price-performance increased? Have agentic tool-using
agents become more efficient? Are we closer to an AI software developer that can efficiently replace humans?
Without data on the resources used, evaluations provide a much less clear picture of current practical
capabilities and the future of AI.

\section*{Acknowledgements}
We thank Zachary Brown for insightful comments on this manuscript.

\bibliographystyle{unsrtnat}
\bibliography{references}

\FloatBarrier   
\clearpage
\onecolumn
\appendix

\section{Data and Analysis Code}\label{appendix:code and data}
Our dataset of benchmark prices, along with the code used for analysis, is available here: 

\href{https://github.com/hansgundlach/Algorithmic_Progress_Inference}{github.com/hansgundlach/Algorithmic\_Progress\_Inference}

\section{Details on Dataset Selection, and Preprocessing}\label{appendix:Details on Dataset Selection, and Preprocessing}
Some models have input and output costs of 0 dollars on Artificial Analysis. We do not include these models in our dataset. We believe these are generally company promotional offers.  Interestingly, in our dataset, we sometimes observe price increases for some models. This is generally due to cheap platforms no longer supporting legacy models. Since we believe that these kinds of price increases are not representative of price-performance decreases, we do not include them in our analysis.
The Internet Archive has only limited data on some models. We collect all information accessible (many Internet Archive pages failed to load). However, if data is logged 6 months apart this could lead to irregular progress estimates.

Epoch benchmarks also include cached tokens. We do not include these for our GPQA-Diamond benchmark cost estimates as the number of cached tokens is generally 20x smaller (as well as around 10x cheaper) than either input or output tokens, and artificial analysis does not have cache token prices. However, for SWE-bench Verified, cached tokens constitute a significant portion of the cost. Therefore, we use proprietary vendors' current cache tokens prices. Vendors generally have a variety of cache token prices for Anthropic models---we use 5m cache write prices. For Deepseek models, we use cache token prices from the official API.  Additionally,  we do not have historical data on cache token prices. However, proprietary models generally do not change either input or output tokens for a given model over time. For instance, we find only one instance of this in our dataset. This is also mentioned by \citet{fradkin2025demand}. Sometimes, Epoch benchmark reports were underspecified, i.e., did not mention which version of a model was used in these cases---we did not include this data. In general, we did not include data in our analysis where we could not match the Epoch model benchmark card with the model name on Artificial Analysis. 

Finally, Epoch includes multiple versions of some models, in particular Claude 3.7 with different reasoning levels. We include these as separate models in our estimates. 

For many models Epoch uses multiple runs (8-16) of a given benchmark i.e runs the benchmark multiple times. We are only able to see the total token numbers so we normalize all benchmarks to 1 iteration by dividing tokens by number of runs. 


\clearpage
\section{Data Tables}

\begin{table*}[h!]
\centering
\caption{Rate of price change across several different benchmarks using general regression approach. Regressions include either all models or only the models that improve in accuracy or price (Pareto Restricted). A separate analysis of only open weight models was possible with GPQA-Diamond (GPQA-Diamond) and OTIS-MOCK AIME 2024-2025 (AIME). Decrease factors $<1$ represent increases.}

\begin{tabular}{|l|l|c|c|c|c|}
\hline
\textbf{Benchmark} & \textbf{Restriction} & \textbf{Annual Reduction Factor} & \textbf{90\% CI} & \textbf{n} & \textbf{$R^2$} \\
\hline

GPQA-Diamond & Pareto Restricted All License & 5.315 & [2.449, 11.534] & 53 & 0.8304 \\ \cline{2-6}
\rowcolor{restrgray} \cellcolor{white} & Pareto Restricted Open Weight & 4.602 & [2.195, 9.648] & 35 & 0.7272 \\ \cline{2-6}
\cellcolor{white} & All License (no restriction)   & 3.769 & [2.085, 6.814] & 135 & 0.6571 \\ \cline{2-6}
\rowcolor{restrgray} \cellcolor{white} & Open Weight (no restriction)  & 1.214 & [0.607, 2.426]  & 72 & 0.5166 \\
\hline

AIME & Pareto Restricted All License & 11.664 & [5.250, 25.911] & 42 & 0.7846 \\ \cline{2-6}
\rowcolor{restrgray} \cellcolor{white} & Pareto Restricted Open Weight & 4.680  & [1.943, 11.273] & 30 & 0.7539 \\ \cline{2-6}
\cellcolor{white} & All License (no restriction)   & 6.988  & [3.687, 13.243] & 109 & 0.6287 \\ \cline{2-6}
\rowcolor{restrgray} \cellcolor{white} & Open Weight (no restriction)  & 2.661  & [1.373, 5.157] & 55 & 0.7087 \\
\hline

SWE\mbox{-}V & Pareto Restricted All License & 4.675 & [0.680, 32.156] & 13 & 0.6281 \\ \cline{2-6}
\rowcolor{restrgray} \cellcolor{white} & Pareto Restricted Open Weight & ---  & [---, ---]      & --- & ---    \\ \cline{2-6}
\cellcolor{white} & All License (no restriction)   & 1.603 & [0.373, 6.881] & 21 & 0.1628 \\ \cline{2-6}
\rowcolor{restrgray} \cellcolor{white} & Open Weight (no restriction)  & ---  & [---, ---]      & --- & ---    \\
\hline

\end{tabular}

\label{tab:regression_results}
\end{table*}

\FloatBarrier
\subsection{Adjusting for GPU Price-Performance Gains}
Trends in benchmark price-performance are also influenced by hardware performance trends. If we want to isolate the component due purely to algorithmic advances, we have to divide the annual factor decrease by the annual hardware price-efficiency gain. Here we use our general regression approach and estimates from \citet{epoch2024priceperformancehardware}, which finds that for a fixed performance level, costs have dropped by $30\%$ a year.

\renewcommand{\arraystretch}{1.15}
\begin{table}[h!]
\centering
\caption{Annual reduction factor (hardware-adjusted) and 90\% CI (hardware-adjusted).}
\label{tab:regression_results_adjusted}
\small
\begin{tabular}{|l|l|c|c|c|c|}
\hline
\textbf{Benchmark} & \textbf{Restriction} &
\shortstack{\textbf{Annual reduction factor}\\\textbf{(hardware-adjusted)}} &
\shortstack{\textbf{90\% CI}\\\textbf{(hardware-adjusted)}} &
\textbf{n} & \textbf{$R^2$} \\
\hline

GPQA-Diamond & Pareto Restricted All License & 3.720 & [1.714, 8.074] & 53 & 0.8304 \\ \cline{2-6}
\rowcolor{restrgray} \cellcolor{white} & Pareto Restricted Open Weight & 3.221 & [1.536, 6.754] & 35 & 0.7272 \\ \cline{2-6}
\cellcolor{white} & All License (no restriction) & 2.638 & [1.459, 4.770] & 135 & 0.6571 \\ \cline{2-6}
\rowcolor{restrgray} \cellcolor{white} & Open Weight (no restriction) & 0.850 & [0.425, 1.698] & 72 & 0.5166 \\
\hline

AIME & Pareto Restricted All License & 8.165 & [3.675, 18.138] & 42 & 0.7846 \\ \cline{2-6}
\rowcolor{restrgray} \cellcolor{white} & Pareto Restricted Open Weight & 3.276 & [1.360, 7.891] & 30 & 0.7539 \\ \cline{2-6}
\cellcolor{white} & All License (no restriction) & 4.891 & [2.581, 9.270] & 109 & 0.6287 \\ \cline{2-6}
\rowcolor{restrgray} \cellcolor{white} & Open Weight (no restriction) & 1.863 & [0.961, 3.610] & 55 & 0.7087 \\
\hline

SWE\mbox{-}V & Pareto Restricted All License & 3.273 & [0.476, 22.509] & 13 & 0.6281 \\ \cline{2-6}
\rowcolor{restrgray} \cellcolor{white} & Pareto Restricted Open Weight & --- & [---, ---] & --- & --- \\ \cline{2-6}
\cellcolor{white} & All License (no restriction) & 1.122 & [0.261, 4.817] & 21 & 0.1628 \\ \cline{2-6}
\rowcolor{restrgray} \cellcolor{white} & Open Weight (no restriction) & --- & [---, ---] & --- & --- \\
\hline
\end{tabular}
\end{table}

\FloatBarrier

\renewcommand{\arraystretch}{1.2}

\begin{table*}[t]
\centering
\caption{Here we show the logit performance trend for all models as well as Pareto and Frontier models (as defined in the paper) with and without price controls.}
\label{tab:multi_benchmark_regression_rounded}
\renewcommand{\arraystretch}{1.2}

\begin{tabular}{|l|l|c|c|c|c|c|}
\hline
\textbf{Benchmark} &
\textbf{Sample} &
\textbf{n} &
\shortstack{\textbf{Time}\\\textbf{Coef}} &
\shortstack{\textbf{Time}\\\textbf{SE}} &
\shortstack{\textbf{Price}\\\textbf{Coef}} &
\textbf{$R^2$} \\
\hline

GPQA-Diamond & Pareto, Without Price Control & 53  & 1.30 & 0.220 & ---        & 0.41 \\ \cline{2-7}
\rowcolor{restrgray} \cellcolor{white} & Pareto, With Price Control    & 53  & 0.68 & 0.055 & 0.570      & 0.88 \\ \cline{2-7}
\cellcolor{white}    & Frontier, Without Price Control & 7 & 1.00 & 0.076 & ---        & 0.97 \\ \cline{2-7}
\rowcolor{restrgray} \cellcolor{white} & Frontier, With Price Control  & 7   & 1.00 & 0.100 & $-0.021$   & 0.97 \\ \cline{2-7}
\cellcolor{white}    & All, Without Price Control    & 138 & 0.96 & 0.130 & ---        & 0.29 \\ \cline{2-7}
\rowcolor{restrgray} \cellcolor{white} & All, With Price Control       & 138 & 0.59 & 0.039 & 0.500      & 0.74 \\
\hline

SWE-Bench & Pareto, Without Price Control & 13 & 1.10 & 0.570 & ---        & 0.24 \\ \cline{2-7}
\rowcolor{restrgray} \cellcolor{white} & Pareto, With Price Control    & 13 & 0.89 & 0.330 & 0.730      & 0.71 \\ \cline{2-7}
\cellcolor{white}    & Frontier, Without Price Control & 5 & 1.00 & 0.200 & ---        & 0.90 \\ \cline{2-7}
\rowcolor{restrgray} \cellcolor{white} & Frontier, With Price Control  & 5  & 0.77 & 0.140 & 0.300      & 0.98 \\ \cline{2-7}
\cellcolor{white}    & All, Without Price Control    & 21 & 0.83 & 0.500 & ---        & 0.13 \\ \cline{2-7}
\rowcolor{restrgray} \cellcolor{white} & All, With Price Control       & 21 & 0.81 & 0.430 & 0.550      & 0.27 \\
\hline

AIME & Pareto, Without Price Control & 42 & 2.60 & 0.630 & ---        & 0.29 \\ \cline{2-7}
\rowcolor{restrgray} \cellcolor{white} & Pareto, With Price Control    & 42 & 2.20 & 0.210 & 1.500      & 0.85 \\ \cline{2-7}
\cellcolor{white}    & Frontier, Without Price Control & 5 & 2.50 & 0.520 & ---        & 0.89 \\ \cline{2-7}
\rowcolor{restrgray} \cellcolor{white} & Frontier, With Price Control  & 5  & 3.30 & 0.790 & $-1.500$   & 0.93 \\ \cline{2-7}
\cellcolor{white}    & All, Without Price Control    & 109 & 2.60 & 0.340 & ---        & 0.34 \\ \cline{2-7}
\rowcolor{restrgray} \cellcolor{white} & All, With Price Control       & 109 & 2.00 & 0.120 & 1.200      & 0.75 \\
\hline
\end{tabular}
\end{table*}

\FloatBarrier
\section{Progress Under Budget Constraints for SWE-bench Verified, and AIME}\label{appendix:added_budget_graphs}

\begin{figure}[h]
  \centering
  \begin{minipage}{0.48\textwidth}
    \centering
    \includegraphics[width=\linewidth,height=0.33\textheight,keepaspectratio]{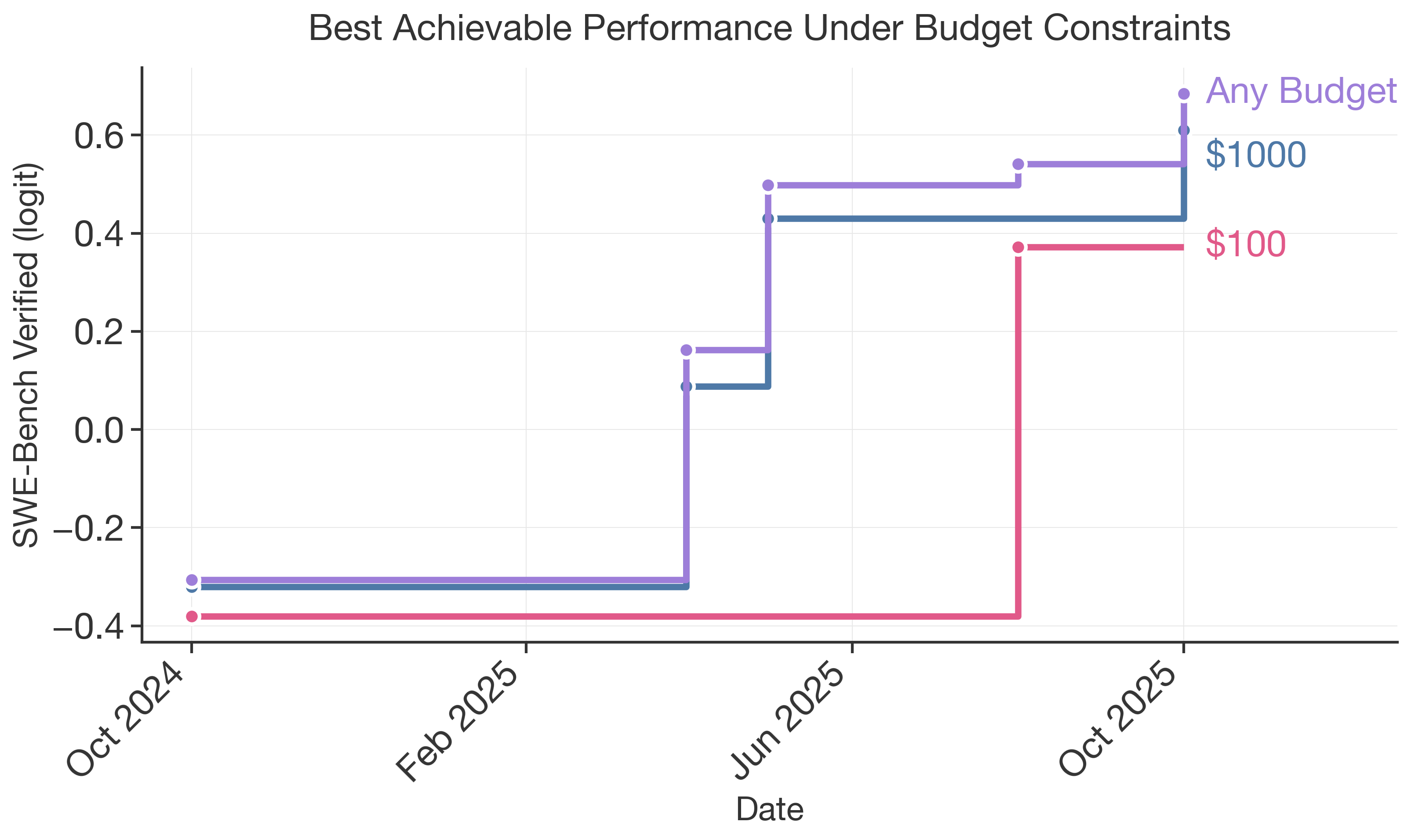}
    \caption{Graph of SWE-bench Verified progress for models under a given budget constraint.}
    \label{fig:swe_budget}
  \end{minipage}
  \hfill
  \begin{minipage}{0.48\textwidth}
    \centering
    \includegraphics[width=\linewidth,height=0.33\textheight,keepaspectratio]{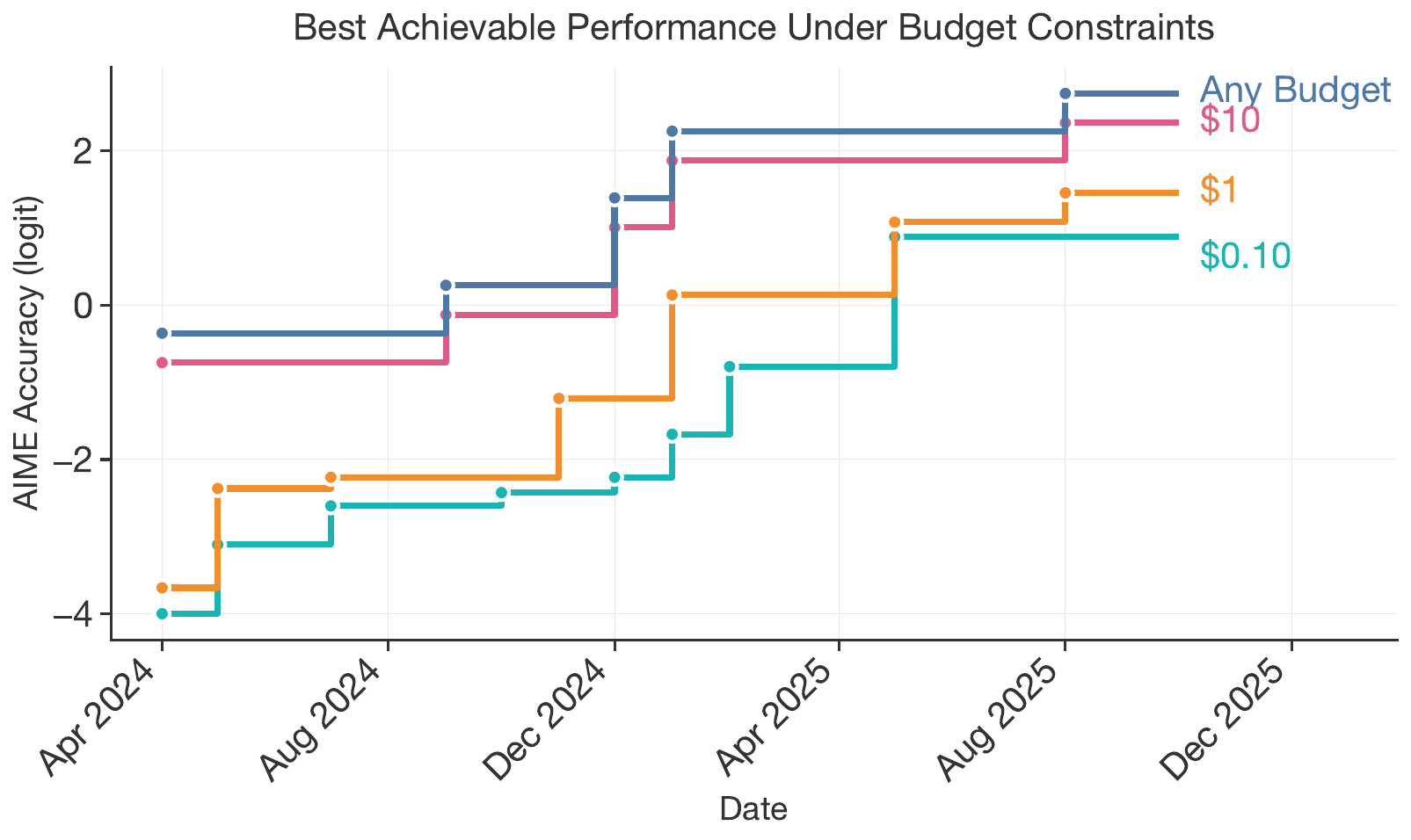}
    \caption{Graph of AIME progress for models under a given budget constraint.}
    \label{fig:aime_budget}
  \end{minipage}
\end{figure}

In Figs.~\ref{fig:swe_budget} and~\ref{fig:aime_budget}, we examine performance on SWE-bench Verified and AIME under progressively tighter budget constraints. Overall, budget has a comparatively modest effect on measured progress. Instead, improvements over time dominate: for example, after roughly three months, a \$0.10 model reaches about the same AIME performance as a \$1.00 model achieved three months earlier.

\end{document}